\documentclass{article}
\usepackage{ijcai26}

\usepackage{times}
\usepackage{soul}
\usepackage{url}
\usepackage[hidelinks]{hyperref}
\usepackage[utf8]{inputenc}
\usepackage[small]{caption}
\usepackage{graphicx}
\usepackage{amsmath}
\usepackage{amsthm}
\usepackage{multirow}
\usepackage{adjustbox}
\usepackage{amssymb}
\usepackage{booktabs}
\usepackage{algorithm}
\usepackage{algorithmic}
\usepackage{subcaption} 
\usepackage{dsfont}
\usepackage[switch]{lineno}

\title{TabKD: Tabular Knowledge Distillation Through Interaction Diversity of Learned Feature Bins}

\author{
Shovon Niverd Pereira
\and
Krishna Khadka\and
Yu Lei\\
\affiliations
Department of Computer Science and Engineering, The University of Texas at Arlington\\
\emails
\{snp3941, krishna.khadka\}@mavs.uta.edu,
ylei@cse.uta.edu,
}

\begin{document}

\maketitle

\begin{abstract}

Data-free knowledge distillation enables model distillation without original training data, which is critical for privacy-sensitive tabular domains. However, existing methods do not perform well on tabular data because they do not explicitly address \emph{feature interactions}, that are critical for encoding predictive knowledge. We identify \textbf{interaction diversity}, systematic coverage of feature combinations, as an important factor for effective tabular model distillation. To operationalize this insight, we propose \textbf{TabKD}, which learns adaptive feature bins aligned with teacher decision boundaries, then generates synthetic queries that ensure uniform pairwise interaction coverage.  Across 4 benchmark datasets and 4 teacher architectures, \textbf{TabKD} achieves highest student-teacher agreement in 14 of 16 configurations, outperforming 5 state-of-the-art baselines. We further show that interaction coverage strongly correlates with distillation quality, validating our core hypothesis.
Our work establishes interaction-focused exploration as a principled framework for tabular model distillation.
\end{abstract}

\section{Introduction}

Data-free knowledge distillation trains a student model to mimic a teacher model's output without the original training data. In many real-world scenarios, organizations possess high-performing tabular models but cannot share the underlying training data. A hospital may deploy a diagnostic model trained on sensitive patient records; a bank may operate a credit scoring system built from proprietary transaction histories. Data-free distillation can be useful in scenarios where an organization wants to train another model with only black-box access to the original model and without the original data. One of the use cases of knowledge distillation is to train a smaller student model for edge devices. However, conventional distillation~\cite{hinton-vinyals:kdnn} still requires representative input samples to query the teacher. This constraint has motivated data-free knowledge distillation (DFKD), where the student learns only from synthetic queries~\cite{fang2019data,chen2019data,lopes2017data,truong2021data}.

Tabular models are fundamentally different from their vision counterparts. While image classifiers detect local patterns through convolutions, tabular models succeed by learning which feature combinations predict outcomes. A credit risk model does not independently assess income or debt; it learns that the specific \emph{combination} of these features signals risk, and neither feature alone suffices to predict the outcome. Unlike image data, where feature interactions are hierarchical and locally correlated, tabular data relies on sharp, non-linear interactions across heterogeneous features with no predefined structure. This fundamental difference explains why existing DFKD methods are developed mainly for vision tasks and cannot effectively capture feature interactions in tabular data. Vision methods rely on spatial inductive biases (convolutions, locality) that have no analog in tabular domains where features interact arbitrarily~\cite{kang2025learning}.

Early work like StealML~\cite{tramer2016stealing} laid the foundation for black-box distillation. Subsequent work ,such as DualCF~\cite{wang2022dualcf} uses counterfactual explanations for binary classification; Marich~\cite{karmakar2023marich} improves query efficiency through active sampling; TabExtractor~\cite{jin2025more} handles feature heterogeneity through entropy-guided generation. Despite these advances, a fundamental bottleneck remains: existing approaches frequently exhibit mode collapse, failing to explore the full decision manifold, and resulting in students that miss critical decision rules~\cite{shin2024teacher}.

We hypothesize that effective tabular model distillation requires systematically exploring the interaction space rather than individual features. This insight is inspired by $t$-way combinatorial testing~\cite{kuhn2013introduction}, a software engineering strategy where covering all $t$-way parameter interactions with minimal samples suffices to expose system behaviors. The key observation is that significant behaviors of complex systems are rarely triggered by all parameters simultaneously, but rather by interactions among a small subset. As in \cite{khadka2024assessing,khadka2026combinatorial}, tabular model decisions similarly depend on interactions among small feature subsets. However, combinatorial testing tries to cover every interaction at least once. This is typically sufficient for fault detection, but not for knowledge distillation. This is because the latter aims to learn the decision patterns of the teacher model. Learning a decision pattern requires observing how the teacher behaves across multiple related samples. Our approach therefore, attempts to have uniform pairwise interaction distributions, ensuring that each decision pattern receives equal representation during training rather than merely guaranteeing minimum coverage. Consider a credit risk model that learns the rule $(\text{age} > 50 \land \text{debt-to-income} < 0.3) \rightarrow \text{low risk}$. Covering all pairwise feature interactions enables the student to observe different decision patterns in a uniform manner.

Our approach is designed to generate samples that achieve interaction diversity. Specifically, the generated samples will cover the space of pairwise interactions in a uniform manner, where each interaction is typically covered multiple times by different samples. Such uniform coverage of pairwise interactions can help expose diverse decision patterns, including those involving higher-order interactions. To operationalize this, we propose \textbf{TabKD}, which discretizes each feature into $K$ learned bins aligned with the teacher's decision boundaries, reducing the infinite input space to a finite set of pairwise interactions that can be systematically covered. Our framework operates in two stages. First, we learn adaptive bin boundaries that partition each feature into semantically meaningful regions where teacher predictions are consistent. Second, we train a generator with an interaction diversity loss that maximizes entropy over pairwise bin combinations, pushing toward uniform coverage rather than mode-collapsing into limited regions. Random sampling may never query specific rare interactions, and entropy-based methods may locate decision boundaries but cannot certify which interactions have been tested. TabKD provides both systematic coverage and boundary-aware sampling.

Experiments across 4 benchmark datasets and 4 teacher architectures (neural networks, XGBoost, Random Forest, TabTransformer) demonstrate that TabKD achieves highest student-teacher agreement in 14 of 16 configurations, outperforming 5 state-of-the-art baselines. We further show that interaction coverage strongly correlates with distillation quality, validating our core hypothesis. Our contributions are:

\begin{enumerate}
    \item \textbf{Approach:} We propose TabKD, a data-free knowledge distillation framework built on the insight that \emph{interaction diversity}, the systematic coverage of feature combinations, is essential for effective tabular distillation. TabKD operationalizes this through dynamic bin learning aligned with teacher decision boundaries and a diversity loss that ensures uniform pairwise interaction coverage.
    \item \textbf{Evaluation:} We perform comprehensive experiments across 4 datasets and 4 teacher architectures, demonstrating that TabKD achieves highest agreement in 14 of 16 configurations over 5 baselines, and that interaction coverage strongly correlates with distillation quality.
    \item \textbf{Tool:} We release our implementation as an open-source tool.\footnote{Hosted at \url{https://github.com/PereiraMavs/int_div}}

\end{enumerate}

\section{Related Work}

Model extraction and distillation research spans query-based methods using auxiliary data and data-free approaches using generators. While knowledge distillation compresses a teachers knowledge into a smaller student, model extraction seeks to steal a model using similar techniques. Both are predominantly developed for vision domains and poorly suited to the heterogeneous feature interactions of tabular data. Related works in this domain are discussed below.

\subsection{Model Extraction}

\paragraph{Query-Based Extraction.} Model extraction attacks~\cite{tramer2016stealing} infer the behavior of a victim model through black-box API queries. Early work focused on equation-solving approaches~\cite{tramer2016stealing}, requiring exponential queries for high-dimensional inputs. Active learning strategies~\cite{papernot2017practical,orekondy2019knockoff} reduce query budgets by selecting informative samples based on uncertainty or gradient approximations. However, these methods assume access to auxiliary unlabeled data from public domain and do not provide any mechanism to assemble most meaningful sample set.

\paragraph{Data-Free Extraction.} Recent work on data-free extraction employs generative models to synthesize training data. DFAD~\cite{fang2019data} uses adversarial training where a generator maximizes student-teacher disagreement. DAFL~\cite{chen2019data} adds activation matching to preserve intermediate representations. These approaches excel in vision domains but may not be sufficient for tabular data due to heterogeneous feature spaces and non-differentiable teachers (tree ensembles). TabExtractor~\cite{jin2025more} addresses non-differentiability through entropy maximization but lacks systematic interaction coverage, leading to mode collapse in high-dimensional feature spaces.

\subsection{Knowledge Distillation}

\paragraph{Standard Distillation.}
Knowledge distillation~\cite{hinton-vinyals:kdnn} transfers knowledge from teacher to student via soft probability labels, preserving "dark knowledge" about class relationships. Variants explore attention transfer~\cite{zagoruyko2016paying}, intermediate representation matching~\cite{romero2014fitnets}, and relational knowledge~\cite{park2019relational}. All assume access to original training data or in-distribution samples.

\paragraph{Data-Free Distillation.}
DFKD methods synthesize training data through generative models. Zero-Shot KD~\cite{wang2021zero} uses metadata and soft label smoothing. DeepInversion~\cite{yin2020dreaming} inverts batch normalization statistics for vision models. CMI~\cite{fang2021contrastive} uses contrastive learning for mode diversity. These methods rely on vision-specific inductive biases (spatial locality, batch norm statistics) absent in tabular domains.

\subsection{Tabular Data Synthesis}

Generative models for tabular data face unique challenges due to mixed feature types and complex dependencies. Among existing works the most prominent few are, CTGAN~\cite{xu2019modeling}, which uses mode-specific normalization and conditional generators. TVAE~\cite{ishfaq2018tvae} employs variational autoencoders with tailored loss functions. However, these methods optimize data realism, not teacher fidelity—generating plausible samples that may not expose student weaknesses or cover decision boundaries.

\paragraph{Gap in Prior Work.}
No existing work addresses the fundamental challenge of systematic interaction coverage in data-free tabular model extraction. Vision-domain methods ignore heterogeneity and interactions; tabular synthesis methods optimize realism over informativeness; active learning assumes auxiliary data. Our work fills this gap with a principled interaction-focused framework.

\section{Background}

\subsection{Data-free Knowledge Distillation}

In the absence of real data, Data-free Knowledge Distillation(DFKD) methods typically employ an adversarial framework to synthesize training samples. The Generator aims to synthesize inputs that maximize the divergence between the Teacher's and the Student's outputs (i.e., "hard" samples), while the Student tries to minimize this divergence. This adversarial game forces the Student to align its decision boundaries with the Teacher's, using the synthetic data as a proxy for the unavailable real data. The student does not need to know the real data, it only need to know how the teacher reacts to particular data-points~\cite{lopes2017data,chen2019data,fang2019data}.

\subsection{Combinatorial Testing}
T-way combinatorial testing is grounded in the empirical interaction fault hypothesis, which suggests that significant behaviors in highly complex systems are rarely triggered by interactions of all parameters simultaneously, but rather by interactions among a relatively small number ($t$) of input parameters~\cite{kuhn2004}. Based on this hypothesis, $t$-way testing aims to generate a minimal, efficient set of samples that guarantees coverage for every possible combination of values across any subset of $t$ parameters. The notion of interaction diversity is inspired by t-way combinatorial testing. In the context of generative modeling, interaction diversity offers a principled metric to encourage a generator to explore the full interaction space rather than mode-collapsing into limited regions. For example, suppose feature A has two bins: Bin 1 = [1–5] and Bin 2 = [6–10], and feature B has two bins: Bin 1 = [OLD] and Bin 2 = [NEW]. This creates four pairwise bin combinations that need to be covered: (A-Bin1, B-Bin1), (A-Bin1, B-Bin2), (A-Bin2, B-Bin1), and (A-Bin2, B-Bin2). To achieve uniform 2-way coverage, the generator produces samples such that the number of samples that land in each combination is approximately the same.

\section{Approach}

Effective tabular distillation requires the student to see samples covering all meaningful feature interactions, not just random points in the input space. Our approach operationalizes this insight in three stages: (1) learn bins aligned with teacher decision boundaries, (2) generate samples with uniform pairwise bin coverage, and (3) train the student on these diverse, challenging samples.

\textbf{Problem Setup.} Let $T: \mathbb{R}^F \to [0,1]^C$ denote a pre-trained teacher mapping $F$-dimensional inputs to $C$ class probabilities, and $S: \mathbb{R}^F \to [0,1]^C$ a lightweight student. We assume no access to the original training data.

\begin{figure*}[t]
    \centering
    \includegraphics[width=0.7\textwidth]{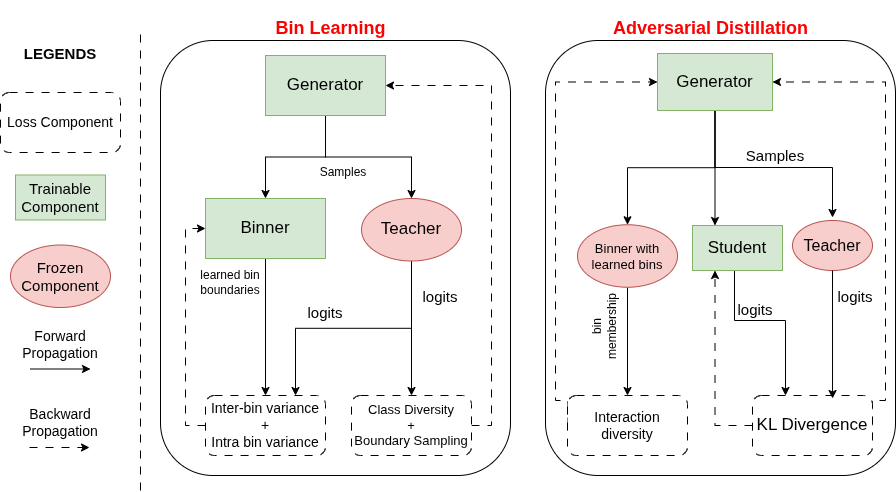}
    \caption{\textbf{TabKD Framework.} The bin learner partitions each feature into semantically meaningful regions based on teacher predictions. The generator then produces samples maximizing pairwise interaction coverage across these bins, while a hardness objective targets student weaknesses. The student learns from this diverse, challenging synthetic data.}
    \label{fig:framework}
\end{figure*}

\subsection{Approach Overview}

TabKD consists of four components working together (Figure~\ref{fig:framework}):

\begin{itemize}
    \item \textbf{Generator} $G(z)$: Maps noise $z \sim \mathcal{N}(0, I)$ to synthetic samples. Trained to maximize both interaction diversity and student--teacher disagreement.

    \item \textbf{Bin Learner} $B(x)$: Learns $K$ adaptive boundaries per feature, partitioning each into regions where teacher predictions are consistent. Outputs soft membership vectors $m \in [0,1]^K$.

    \item \textbf{Teacher} $T(x)$: Teacher is a target frozen pre-trained model providing ground-truth soft labels.

    \item \textbf{Student} $S(x)$: Student model is a smaller network trained to mimic the teacher. See section 5.3
\end{itemize}

Training proceeds in three phases to ensure stable convergence:

\begin{enumerate}
    \item \textbf{Warmup:} Pre-train student on uniform random samples; populate replay buffer (used later to prevent catastrophic forgetting during adversarial training~\cite{mnih2015human}).

    \item \textbf{Bin Learning:} Train the bin learner to find decision-boundary-aligned discretizations; freeze upon convergence.

    \item \textbf{Adversarial Distillation:} Generator and student alternate, generator maximizes interaction diversity and hardness; student minimizes divergence from teacher.
\end{enumerate}

The staged approach prevents co-adaptation between generator and bin learner, ensuring coverage reflects genuine feature space exploration. The following subsections detail each component.

\subsection{Dynamic Bin Learning}

To achieve interaction diversity, we first need a effective discretization of each feature. Random or uniform bins waste coverage on homogeneous regions where the teacher's predictions are constant. Instead, we learn bins aligned with the teacher's decision boundaries. We aim to ensure that covering all bin combinations captures the majority of behaviorally distinct regions.

\textbf{Intuition.} The teacher's decision boundary implicitly partitions each feature into regions of consistent predictions. Consider a feature from a dataset where the boundary crosses once, separating values $[1\text{--}5]$ (Class~0) from $[6\text{--}10]$ (Class~1). These boundary crossings define \emph{natural} bin boundaries. By learning bins aligned with these transitions, we ensure that covering all bin combinations means covering all behaviorally distinct regions.

\textbf{Learning Objective.} We train the bin learner to minimize intra-bin prediction variance (ensuring homogeneity within bins) while maximizing inter-bin prediction variance (ensuring bins capture different behaviors):
\begin{equation}
\mathcal{L}_{\text{bin}} = \lambda_{\text{intra}} \cdot \text{Var}_{\text{intra}}(M, P_T) + \lambda_{\text{inter}} \cdot \frac{1}{\text{Var}_{\text{inter}}(M, P_T)}
\end{equation}
where $M$ denotes soft bin memberships and $P_T$ the teacher's predictions. Soft membership is required to ensure the assignment is differentiable, which a an discrete assignment cannot provide~\cite{frosst2017distillingneuralnetworksoft}. The variance is calculated as the spread of probability of samples inside each bin weighted with soft membership for each sample.

\textbf{Boundary-Focused Sampling.} Learning effective bins requires samples near decision boundaries, where predictions transition between classes. During bin learning, we train a a generator to target these regions:
\begin{equation}
\mathcal{L}_{\text{gen}}^{(1)} = \lambda_{\text{div}} \cdot \mathcal{L}_{\text{class-div}}(S(X_{\text{gen}})) + \lambda_{\text{boundary}} \cdot \mathcal{L}_{\text{entropy}}(T(X_{\text{gen}}))
\end{equation}
where $\mathcal{L}_{\text{class-div}}$ ensures class balance and $\mathcal{L}_{\text{entropy}}$ targets high-uncertainty regions via entropy maximization.

\textbf{Why Staged Training?} Jointly optimizing bins and coverage causes instability: the generator fills bins faster than boundaries stabilize. Our staged approach, freezing bins before adversarial training, prevents this co-adaptation. With bins frozen, we can now define systematic coverage.

\subsection{Interaction Diversity Loss}

This is the core contribution of TabKD: a loss function that ensures systematic coverage of feature combinations. With frozen bins providing a effective discretization, we can now formalize what it means to ``cover'' the interaction space.

\textbf{Formalization.} For pairwise ($t=2$) coverage, we consider all $\binom{F}{2}$ feature pairs. For each pair $(i, j)$, the joint bin assignment forms a $K \times K$ matrix. Ideally, generated samples should uniformly populate all $K^2$ cells, ensuring every combination of feature $i$'s regions with feature $j$'s regions is explored.

We compute the empirical joint distribution over bin combinations:
\begin{equation}
\label{eq:3}
P(k_1, k_2 \mid i, j) = \frac{1}{N} \sum_{x} m_{k_1}^{(i)}(x) \cdot m_{k_2}^{(j)}(x)
\end{equation}
where $N$ is batch size and $m_{k}^{(i)}(x)$ is the soft membership of sample $x$ in bin $k$ of feature $i$.

\textbf{Diversity Loss.} Maximum entropy over this distribution corresponds to uniform coverage. Our diversity loss maximizes this entropy:
\begin{equation}
\mathcal{L}_{\text{diversity}} = -\frac{1}{\binom{F}{2}} \sum_{i < j} H(P(\cdot, \cdot \mid i, j))
\end{equation}

By minimizing $\mathcal{L}_{\text{diversity}}$ (negative entropy), we push the generator toward uniform pairwise coverage, systematically exploring the interaction space rather than mode-collapsing.

\subsection{Adversarial Sample Generation}

Diversity alone is insufficient, we also want samples that challenge the student's current approximation. We combine interaction coverage with a hardness objective.

\textbf{Hardness Loss.} We measure student--teacher disagreement via KL divergence:
\begin{equation}
\mathcal{L}_{\text{hardness}} = -\frac{1}{N} \sum_{x} D_{\text{KL}}(T(x) \parallel S(x))
\end{equation}
The negative sign means the generator is rewarded for finding samples where the student fails. KL divergence naturally upweights uncertain regions near decision boundaries, where $T(x) \approx 0.5$.

\textbf{Combined Objective.} The generator minimizes:
\begin{equation}
\label{eq:6}
\mathcal{L}_{\text{gen}}^{(2)} = \mathcal{L}_{\text{diversity}} + \lambda_{\text{hard}} \cdot \mathcal{L}_{\text{hardness}}
\end{equation}

This balances systematic exploration (covering all interactions, $\mathcal{L}_{\text{diversity}}$) with targeted exploitation (finding student weaknesses, $\mathcal{L}_{\text{hardness}}$). The diversity term encourages interaction coverage; the hardness term ensures samples are informative for learning. $\mathcal{L}_{\text{gen}}^{(2)}$ in equation \ref{eq:6}, used as the loss to the generator during adversarial distillation

\subsection{Student Training}

The student minimizes KL divergence from the teacher's soft predictions:
\begin{equation}
\mathcal{L}_{\text{student}} = \frac{1}{N} \sum_{x} D_{\text{KL}}(S(x) \parallel T(x))
\end{equation}

Training exclusively on adversarial samples risks forgetting previously learned knowledge. To mitigate this, we maintain a replay buffer populated during warmup with random samples. Each update uses 90\% adversarial samples and 10\% replay samples, balancing challenge with stability.

\textbf{Summary.} We have explained three key mechanisms: (1) dynamic bin learning to discretize features meaningfully, (2) interaction diversity loss to ensure systematic coverage, and (3) hardness-guided generation to target student weaknesses. Together, these operationalize our core insight that interaction diversity can play an important role for effective tabular model distillation.

\section{Experimental Setup}

\subsection{Research Question}
Our primary research questions are:

\textbf{RQ1: Agreement Accuracy.} How accurately does TabKD replicate teacher model behavior, as measured by teacher--student agreement and test accuracy?

\textbf{RQ2: Comparison.} How does TabKD compare against state-of-the-art data-free distillation methods in extraction quality?

\textbf{RQ3: Interaction Diversity vs Agreement.} Does systematic feature interaction coverage correlate with improved distillation quality?

\subsection{Datasets}

We evaluate our framework on four benchmark classification datasets, previously used by multiple works ~\cite{jin2025more,ma2023divtheft,arik2021tabnet,gao2024sel,hollmann2022tabpfn}.
\begin{itemize}
    \item Adult~\cite{adult_2}: This dataset contains 48K samples with 14 features and predicts whether an individual's income exceeds \$50K.
    \item Breast Cancer~\cite{breast_cancer_14}: This dataset contains 569 samples with 30 features derived from cell nuclei measurements to classify tumors as malignant or benign.
    \item Credit~\cite{default_of_credit_card_clients_350}: This dataset contains 30K samples with 23 features and predicts credit card default.
    \item Mushroom~\cite{mushroom_73}: This dataset contains 8K samples with 22 features describing physical characteristics to classify mushrooms as edible or poisonous.
\end{itemize}
All datasets are binary classification and features are normalized using standard scaling, and we use an 80/20 train-test split with stratified sampling.

\subsection{Model Architectures}

\textbf{Teacher Models.} We consider four teacher architectures spanning neural, ensemble, and transformer-based methods:
\begin{itemize}
    \item Neural Network (NN): The neural network consists of two hidden layers ($128 \to 64$ units) with ReLU activations and dropout ($p = 0.2$).
    \item XGBoost~\cite{chen2016xgboost}: The gradient boosting model uses 100 estimators with maximum depth 6 and learning rate 0.1.
    \item Random Forest~\cite{breiman2001random}: The random forest comprises 100 trees with maximum depth 10 and minimum samples per split of 5.
    \item TabTransformer~\cite{huang2020tabtransformer}: The transformer-based model with multi-head self-attention for feature encoding in tabular data.
\end{itemize}

\textbf{Student Model.} A single-hidden-layer network (32 units, ReLU activation) that provides a uniform, lightweight approximation regardless of the teacher's architecture. Whether the teacher is a neural network, tree ensemble, or transformer, the student offers a consistent inference pipeline with predictable latency and differentiability, making it suitable for deployment on resource-constrained devices.

\subsection{Baselines}
We compare against five baselines representing distinct extraction paradigms: (1)~\textbf{StealML}~\cite{tramer2016stealing}, the seminal sampling-based attack with adaptive boundary refinement; (2)~\textbf{TabExtractor}~\cite{jin2025more}, a data-free method using entropy-guided synthetic data generation; (3)~\textbf{CF}~\cite{aivodji2020model} and (4)~\textbf{DualCF}~\cite{wang2022dualcf}, which leverage counterfactual explanations for extraction; and (5)~\textbf{DivT}~\cite{ma2023divtheft}, an active learning approach employing divide-and-conquer query selection.

\paragraph{Implementation.}
We use official implementations for StealML and CF where available. For TabExtractor, DualCF, and DivT, we reimplemented them following the methodology described in the original papers, as no official code was released. Following  TabExtractor~\cite{jin2025more}, all baselines use a query budget of 9,600 samples.

\subsection{Training Protocol}

Training proceeds in three phases. In Phase~0 (Warmup), the student is pre-trained for 30 epochs(means update steps with each batch) on randomly sampled instances using standard KL-divergence distillation:
\begin{equation}
\mathcal{L}_{\text{warmup}} = D_{\text{KL}}\bigl(P_{\text{student}}(x) \parallel P_{\text{teacher}}(x)\bigr).
\end{equation}

Phase~1 (Boundary Stabilization) spans 200 epochs during which the adaptive bin boundaries $b_{f,k}$ are learned. We apply linear temperature annealing:
\begin{equation}
\tau(e) = \tau_{\text{start}} - \frac{(\tau_{\text{start}} - \tau_{\text{end}}) \cdot e}{E_{\text{phase1}}}.
\end{equation}

In Phase~2 (Adversarial Distillation), training continues for 400 epochs with fixed bin boundaries, focusing on adversarial sample generation to maximize coverage of the input space.

\subsection{Temperature Schedules}

Ensemble teachers produce sharper probability distributions than neural networks, necessitating teacher-specific temperature schedules to preserve soft label information during distillation. Table~\ref{tab:modeltemp} summarizes these configurations.

\begin{table}[htbp]
\centering
\renewcommand{\arraystretch}{0.50}
\begin{tabular}{lcccc}
\toprule
Teacher & $\tau_{\text{start}}$ & $\tau_{\text{end}}$ & $\tau_{\text{phase2}}$ & $T_{\text{distill}}$ \\
\midrule
Neural Network & 1.0 & 0.05 & 0.2 & 1.0 \\
Random Forest & 1.2 & 0.08 & 0.25 & 1.5 \\
XGBoost & 1.5 & 0.10 & 0.4 & 2.0 \\
TabTransformer & 1.2 & 0.08 & 0.25 & 1.2\\
\bottomrule
\end{tabular}
\caption{Teacher-specific temperature schedules. Higher temperatures for ensemble methods compensate for their typically sharper output distributions.}
\label{tab:modeltemp}
\end{table}

\subsection{Hyperparameters}

We use a batch size of 128 and optimize with Adam (learning rate 0.001) with cosine annealing. The coverage and hard-label loss weights are set to $\lambda_{\text{cov}} = 10.0$ and $\lambda_{\text{hard}} = 2.0$, respectively. We discretize each feature into $K = 8$ bins.

\subsection{Evaluation Metrics}

We report four metrics to assess distillation quality and input space exploration.

\textbf{Test Accuracy} measures the proportion of correctly predicted instances on held-out test data.

\textbf{F1 Score} is the harmonic mean of precision and recall.

\textbf{Teacher-Student Agreement} captures the proportion of test instances where the student and teacher produce identical predictions.

\textbf{Cumulative Coverage} quantifies the fraction of pairwise feature-bin interactions visited during training.

\noindent These metrics are formally defined as:
\begin{align}
    \text{Accuracy} &= \frac{TP + TN}{TP + TN + FP + FN}, \\
    \text{F1} &= \frac{2 \cdot \text{Precision} \cdot \text{Recall}}{\text{Precision} + \text{Recall}}, \\
    \text{Agreement} &= \frac{1}{N} \sum_{i=1}^{N} \mathds{1}\bigl[\hat{y}_i^{(s)} = \hat{y}_i^{(t)}\bigr], \\
    \text{Coverage} &= \frac{|\mathcal{V}|}{\binom{F}{2} \cdot K^2},
\end{align}
where $TP$, $TN$, $FP$, and $FN$ denote true positives, true negatives, false positives, and false negatives, respectively. $N$ is the number of test instances, $\hat{y}_i^{(s)}$ and $\hat{y}_i^{(t)}$ are the student and teacher predictions for instance $i$, $\mathcal{V}$ is the set of visited feature-bin pairs, $F$ is the number of features, and $K$ is the number of bins per feature. The results are the average of 5 runs. 

\textbf{Efficiency:} The parallelizable computation related to pairwise coverage is to compute the joint distribution over bin combinations, in equation \ref{eq:3}, and the cost for N samples is $O(F^2 · K^2 · N )$.

\section{Results}

We evaluated our knowledge distillation framework across four benchmark datasets with four different teacher architectures: Neural Network (NN), XGBoost, Random Forest (RF), and TabTransformer. Table~\ref{tab:results} presents the comprehensive performance metrics, including student-teacher agreement, test accuracy, F1 score, AUC score, and bin coverage.

\begin{table*}[h!]
\centering
\renewcommand{\arraystretch}{0.70}
\begin{adjustbox}{max width=\textwidth}
\begin{tabular}{ll|cccc|cccc|cccc|cccc}
\toprule
& & \multicolumn{4}{c|}{\textbf{Neural Net}} & \multicolumn{4}{c|}{\textbf{XGBoost}} & \multicolumn{4}{c}{\textbf{Random Forest}} & \multicolumn{4}{c}{\textbf{TabTransformer}} \\
\textbf{Dataset} & \textbf{Method} & Acc & F1 & AUC & Agree & Acc & F1 & AUC & Agree & Acc & F1 & AUC & Agree & Acc & F1 & AUC & Agree \\
\midrule
\multirow{6}{*}{Adult} & StealML & 77.4 & 70.9 & 81.4 & 85.4 & 73.3 & 62.7 & 64.7 & 78.6 & 60.6 & 57.1 & 68.4 & 64.2 & 77.7 & 77.9 & 65.9 & 83.3 \\
& TabExtractor & 71.1 & 76.5 & 74.0 & 76.8 & 76.2 & 67.8 & 57.3 & 74.8 & \textbf{78.9} & 68.1 & \textbf{69.2} & 69.1 & 77.2 & 76.8 & 67.3 & 84.9 \\
& CF & 62.2 & 60.1 & 73.8 & 62.5 & 52.9 & 61.3 & 62.9 & 69.4 & 64.9 & 55.9 & 62.2 & 59.3 & 62.2 & 63.4 & 59.9 & 65.1 \\
& DualCF & 74.6 & 69.2 & 80.8 & 80.4 & 53.8 & 69.6 & 53.3 & 53.5 & 57.1 & 54.8 & 67.2 & 54.6 & 70.4 & 65.7 & 64.5 & 62.9 \\
& DivT & 79.8 & 72.1 & \textbf{82.6} & \textbf{91.4} & 62.7 & 51.2 & 53.0 & 66.2 & 68.4 & 67.7 & 62.8 & 52.9 & 78.0 & \textbf{78.9} & 68.5 & 85.0 \\
& TabKD (Ours) & \textbf{80.0} & \textbf{79.3} & 81.3 & 91.0 & \textbf{76.5} & \textbf{77.0} & \textbf{78.0} & \textbf{81.1} & 77.0 & \textbf{70.5} & 62.0 & \textbf{84.4} & \textbf{78.3} & 77.2 & \textbf{76.1} & \textbf{86.1} \\
\midrule
\multirow{6}{*}{Credit} & StealML & 77.8 & 52.9 & 52.0 & 90.9 & 62.9 & 50.8 & 51.5 & 69.5 & 74.0 & 54.2 & 54.0 & 84.0 & 71.3 & 75.6 & \textbf{76.5} & 79.2 \\
& TabExtractor & 77.9 & 58.9 & 52.3 & 90.1 & 69.4 & 57.4 & \textbf{59.2} & 80.1 & 75.4 & 50.4 & 54.1 & 85.1 & \textbf{78.7} & 63.7 & 65.1 & 81.4 \\
& CF & 77.8 & 56.2 & \textbf{75.2} & 89.7 & 76.7 & 54.8 & 66.9 & 85.8 & \textbf{77.3} & \textbf{76.7} & \textbf{77.4} & 87.5 & 66.1 & 59.3 & 66.0 & 74.4 \\
& DualCF & 76.0 & 54.0 & 57.7 & 88.8 & 64.6 & 51.8 & 53.0 & 71.5 & 72.0 & 71.3 & 77.4 & 80.2 & 66.1 & 54.9 & 65.6 & 65.0 \\
& DivT & 77.5 & 51.7 & 55.3 & 90.8 & 72.1 & 58.5 & 68.9 & 80.7 & 62.8 & 56.7 & 45.5 & 67.7 & 77.8 & 67.8 & 66.2 & 80.5 \\
& TabKD (Ours) & \textbf{80.3} & \textbf{76.7} & 65.1 & \textbf{97.0} & \textbf{77.7} & \textbf{68.5} & 58.6 & \textbf{88.0} & 74.5 & 68.0 & 49.0 & \textbf{87.7} & 77.6 & \textbf{69.5} & 52.7 & \textbf{87.1} \\
\midrule
\multirow{6}{*}{Breast Cancer} & StealML & 77.2 & 77.2 & 99.1 & 75.4 & 91.2 & 91.0 & 98.0 & 88.6 & 59.6 & 58.8 & 99.6 & 57.0 & 76.8 & 72.9 & 76.2 & 76.0 \\
& TabExtractor & 86.8 & 86.9 & 90.4 & 85.1 & 77.4 & 81.5 & 88.6 & 86.1 & 76.8 & 68.1 & 98.0 & 71.9 & 80.3 & 80.1 & 79.4 & 78.6 \\
& CF & 84.8 & 86.3 & 87.2 & 86.0 & 76.8 & 78.9 & 74.5 & 79.2 & 79.8 & 80.7 & 82.9 & 74.2 & 85.6 & 82.3 & 89.4 & 85.6 \\
& DualCF & 84.2 & 84.0 & 99.1 & 83.3 & 75.4 & 75.4 & 93.9 & 72.8 & 76.5 & 81.2 & 95.4 & 81.2 & 88.2 & 88.1 & 89.8 & 89.1 \\
& DivT & 87.7 & 87.5 & 99.4 & 87.7 & 88.6 & 88.2 & 97.5 & 89.5 & 86.1 & 88.8 & 99.3 & 83.5 & 86.5 & 86.3 & 88.6 & 86.5 \\
& TabKD (Ours) & \textbf{95.6} & \textbf{95.6} & \textbf{99.5} & \textbf{95.6} & \textbf{95.6} & \textbf{93.9} & \textbf{98.5} & \textbf{96.5} & \textbf{93.9} & \textbf{93.9} & \textbf{99.7} & \textbf{91.2} & \textbf{90.4} & \textbf{91.4} & \textbf{98.4} & \textbf{90.4} \\
\midrule
\multirow{6}{*}{Mushroom} & StealML & 73.0 & 67.5 & 78.1 & 94.0 & 70.0 & 54.8 & 74.6 & 79.5 & 71.5 & 55.9 & 76.2 & 85.5 & 82.3 & 83.3 & 83.4 & 84.8 \\
& TabExtractor & 72.5 & 67.4 & 71.4 & 90.5 & 73.0 & 69.2 & 74.7 & 76.5 & 76.0 & 72.0 & 76.7 & 79.0 & 84.3 & 85.1 & 79.2 & 84.6 \\
& CF & 57.0 & 50.6 & 51.0 & 63.0 & 64.5 & 66.5 & 68.3 & 70.0 & 61.5 & 50.9 & 50.1 & 65.5 & 73.9 & 73.9 & 70.0 & 74.1 \\
& DualCF & 74.5 & 67.5 & 77.3 & 90.0 & 70.0 & 58.3 & 73.6 & 80.5 & 73.5 & 57.4 & 78.8 & 88.5 & 68.8 & 66.4 & 69.9 & 67.3 \\
& DivT & 77.0 & 71.2 & 78.5 & 93.0 & 70.0 & 56.3 & 72.0 & 81.5 & 73.0 & 60.0 & 75.8 & 88.0 & \textbf{86.6} & 83.5 & 88.1 & \textbf{88.3} \\
& TabKD (Ours) & \textbf{92.1} & \textbf{92.1} & \textbf{96.8} & \textbf{96.5} & \textbf{83.6} & \textbf{83.5} & \textbf{89.8} & \textbf{83.5} & \textbf{91.1} & \textbf{91.1} & \textbf{94.8} & \textbf{91.1} & 84.2 & \textbf{84.1} & \textbf{90.7} & 85.2 \\
\bottomrule
\end{tabular}
\end{adjustbox}
\caption{Distillation Performance Comparison across Datasets and Teacher Types. Agreement (\%) is the primary metric for extraction quality. \textbf{Bold} indicates best result per column.}
\label{tab:results}
\end{table*}

\begin{figure*}[h]
     \centering
     \begin{subfigure}{0.24\textwidth}
         \centering
         \includegraphics[width=\textwidth]{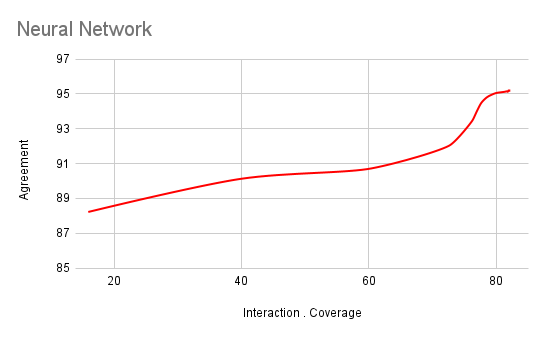}
         \caption{Neural Network Teacher}
         \label{fig:phase1}
     \end{subfigure}
     \begin{subfigure}[b]{0.24\textwidth}
         \centering
         \includegraphics[width=\textwidth]{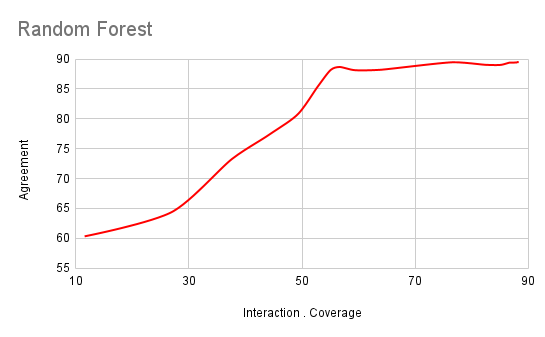}
         \caption{Random Forest Teacher}
         \label{fig:phase2}
     \end{subfigure}
     \begin{subfigure}[b]{0.24\textwidth}
         \centering
         \includegraphics[width=\textwidth]{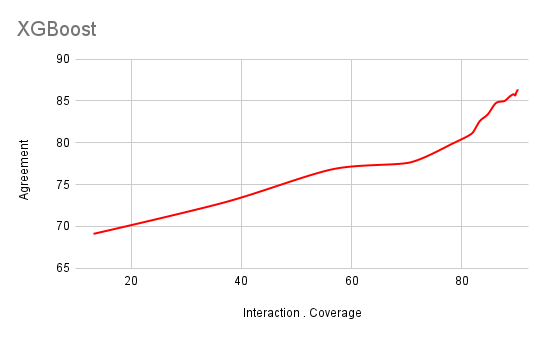}
         \caption{XGBoost Teacher}
         \label{fig:phase3}
     \end{subfigure}
     \begin{subfigure}[b]{0.24\textwidth}
         \centering
         \includegraphics[width=\textwidth]{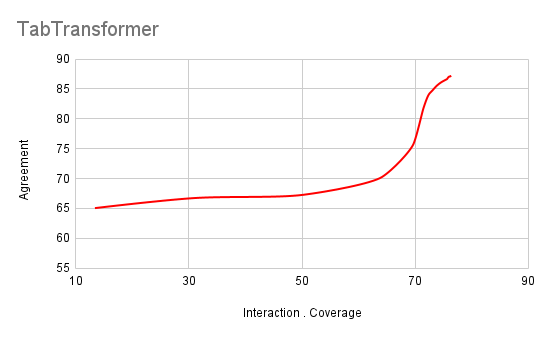}
         \caption{TabTransformer Teacher}
         \label{fig:phase3}
     \end{subfigure}
        \caption{Overview of the Agreement Vs Interaction Coverage.}
        \label{fig:three_phases}
\end{figure*}

\subsection{Overall Performance}

TabKD demonstrates strong and consistent performance across all four teacher architectures. Overall, TabKD achieves the best agreement in 14 out of 16 dataset-teacher combinations, demonstrating robustness across different teacher architectures.

\subsection{RQ1: Agreement Accuracy}
Agreement accuracy is measured by the percentage of samples predicted by the student that match the teacher's predictions. Across all configurations, the Neural Network teacher achieves the highest agreement of 97.0\% on the Credit dataset, while XGBoost has the lowest agreement of 81.1\% on the Adult Income dataset. On average across the four datasets, Neural Network, XGBoost, Random Forest, and TabTransformer teachers achieve agreement rates of 95.0\%, 87.3\%, 88.6\%, and 87.2\% respectively. In phase 3, the use of interaction diversity pushes the generator to produce samples covering the vast majority of 2-way interaction combinations, thus creating diversity. Additionally, the teacher-student disagreement objective provides incentive to create samples that are hard for the student to predict.

\subsection{RQ2: Comparison with Other Baselines}
Compared to five other baselines tested on the same datasets, the overall performance of \textit{TabKD} is consistently better. In terms of agreement accuracy, TabKD achieves the best results in 14 out of 16 dataset-teacher combinations. The exceptions are: (1) Adult with Neural Network, where DivT achieves slightly higher agreement (91.4\% vs 91.0\%), and (2) Mushroom with TabTransformer, where DivT outperforms on both accuracy (86.6\% vs 84.2\%) and agreement (88.3\% vs 85.2\%). For all other combinations, TabKD shows superior agreement accuracy. In terms of model accuracy, \textit{TabKD} also demonstrates strong performance across the board. All results are presented in Table \ref{tab:results}. Current works do not address systematic interaction coverage, which is particularly important for tabular data. As a result, our dynamic binning feature, which creates optimal bins, combined with interaction diversity, achieves better results than existing approaches.

\subsection{RQ3: Interaction Diversity vs Agreement}

One of our main contributions suggests that as \textit{t}-way combinations of features are covered more thoroughly by the generator, we should observe better agreement accuracy. Figure \ref{fig:three_phases} shows that for all three teacher models, average agreement accuracy increases as interaction diversity increases. Because of the well-formed bins and enforced interaction diversity, \textit{TabKD} covers important 2-way interactions, and as coverage increases, the student learns from new samples and agreement improves accordingly.

\section{Ablation Study}

\begin{table}[h]
\centering
\renewcommand{\arraystretch}{0.45}
\resizebox{\columnwidth}{!}{%
\begin{tabular}{@{}lcccc@{}}
\toprule
\textbf{Teacher} & \textbf{Adult} & \textbf{Credit} & \textbf{Cancer} & \textbf{Mushroom} \\
\midrule
\multicolumn{5}{@{}l}{\textit{With Dynamic Binning}} \\
\midrule
NN            & 92.00 & 97.03 & 95.61 & 96.50 \\
XGBoost       & 81.13 & 88.50 & 96.50 & 83.50 \\
Random Forest & 84.41 & 88.20 & 93.90 & 91.08 \\
TabTransformer & 86.1 & 87.1 & 90.4 & 85.2\\
\midrule
\multicolumn{5}{@{}l}{\textit{Without Dynamic Binning (Static)}} \\
\midrule
NN            & 87.00 & 91.00 & 93.00 & 96.00 \\
XGBoost       & 42.00 & 88.00 & 93.00 & 83.30 \\
Random Forest & 58.90 & 88.20 & 88.60 & 91.00 \\
TabTransformer & 33.0 & 86.6 & 90.7 & 87.4\\
\bottomrule
\end{tabular}%
}
\caption{Student Accuracy Comparison: Dynamic vs Static Binning}
\label{tab:ablation_dynamic_vs_static}
\end{table}

One of the important component of \textit{TabKD} is the dynamic binning module. It is important learn the bins to find out optimal boundaries where prediction distribution remains similar. To find out the importance we removed the dynamic binning module and calculated agreement accuracy with preset static bins. The static bins are of equal lengths covering the whole feature length. The results in table \ref{tab:ablation_dynamic_vs_static} shows that for complex and imbalanced datasets like \textit{Adult} shows significant decrement of performance. But for simpler datasets like \textit{Mushroom}, where we have categorical features with clear rules uniform bins perform similar to adjusted bins. 

\section{Conclusion}

We introduced \emph{TabKD}, a data-free knowledge distillation framework for tabular data that leverages feature interaction diversity, inspired by combinatorial testing, to generate informative samples for student models. It uses dynamic bin learning based on prediction similarities of feature ranges. Experiments across four datasets and four teacher architectures show \emph{TabKD} achieves best agreement in 14/16 configurations over five baselines.

While TabKD demonstrates strong performance across diverse settings, two aspects warrant future exploration. First, the number of bins per feature is currently set as a static hyperparameter. It is possible that fewer or more bins may be more optimal for particular features; future work could extend the bin learner to adaptively determine the optimal granularity for each feature. Second, TabKD focuses on 2-way interaction coverage, which proves effective for the evaluated datasets. However, for more complex datasets, higher-order combinations may be beneficial. Exploring whether a particular $t$-way interaction strength improves performance for specific dataset characteristics remains an open direction.

\bibliographystyle{named}
\bibliography{ijcai26}

\end{document}